\documentclass[review]{elsarticle}
\graphicspath{ {./figures/} }
\usepackage{hyperref}
\usepackage{float}
\usepackage{verbatim} 
\usepackage{apalike}
\usepackage{multirow}
\usepackage{tabularx}
\usepackage{amssymb}
\usepackage{placeins}
\restylefloat{figure}
\restylefloat{table}

\journal{}

\biboptions{authoryear}

\begin{document}
\begin{frontmatter}

\begin{titlepage}
\begin{center}
\vspace*{1cm}

\textbf{ \large From Natural Language to Simulations: Applying GPT-3 Codex to Automate Simulation Modeling of Logistics Systems}

\vspace{1.5cm}

Ilya Jackson$^{a}$ (ilyajack@mit.edu), Maria Jesus Saenz$^b$ (mjsaenz@mit.edu) \\

\hspace{10pt}

\begin{flushleft}
\small  
$^a$ MIT Center for Transportation and Logistics. Room E40-357, 1 Amherst St, Cambridge, MA 02139, United States \\
$^b$ MIT Center for Transportation and Logistics. Room E40-379, 1 Amherst St, Cambridge, MA 02139, United States

\vspace{1cm}
\textbf{Corresponding Author:} \\
Ilya Jackson \\
MIT Center for Transportation and Logistics. Room E40-357, 1 Amherst St, Cambridge, MA 02139, United States \\
Tel: (617) 258-7569 \\
Email: ilyajack@mit.edu

\end{flushleft}        
\end{center}
\end{titlepage}

\title{From Natural Language to Simulations: Applying GPT-3 Codex to Automate Simulation Modeling of Logistics Systems}

\author[label1]{Ilya Jackson\corref{cor1}}
\ead{ilyajack@mit.edu}

\author[label1]{Maria Jesus Saenz}
\ead{mjsaenz@mit.edu}

\cortext[cor1]{Corresponding author.}
\address[label1]{MIT Center for Transportation and Logistics. 1 Amherst St, Cambridge, MA 02139, United States}

\begin{abstract}
Our work is the first attempt to apply Natural Language Processing to automate the development of simulation models of systems vitally important for logistics. We demonstrated that the framework built on top of the fine-tuned GPT-3 Codex, a Transformer-based language model, could produce functionally valid simulations of queuing and inventory control systems given the verbal description. In conducted experiments, GPT-3 Codex demonstrated convincing expertise in Python as well as an understanding of the domain-specific vocabulary. As a result, the language model could produce simulations of a single-product inventory-control system and single-server queuing system given the domain-specific context, a detailed description of the process, and a list of variables with the corresponding values. The demonstrated results, along with the rapid improvement of language models, open the door for significant simplification of the workflow behind the simulation models’ development, which will allow experts to focus on the high-level consideration of the problem and holistic thinking.

\end{abstract}
\end{frontmatter}

\section{Introduction}
\label{introduction}
Global supply chains are the “nuclear reactor” of globalization and the cornerstone of the modern economy \citep{sheffi_new_2020}. From the semiconductor plant in Hsinchu to the advanced automotive factory in Fremont, from the smart pharmaceutical manufacturing in Freiburg to Cameroonian cocoa farms, every economic actor faces the problem of transporting the right commodities to the right place at the right time. Logistics systems constitute the backbone of global supply chains and play a strategic role in the worldwide economy by propagating material flows through increasingly complex, interconnected, and multimodal transportation channels \citep{schenker_futureproof_2019}.

The global interconnectedness and complexity go hand in hand with the lack of visibility, uncertainty, and constant pressure from consumer-driven markets. The latter fact forces businesses to make critical decisions quickly and confidently. Therefore, the need for modeling logistics systems and associated processes is self-evident \citep{ivanov_viability_2020}. Quantitative models can facilitate such core activities as forecasting, planning, scenario analysis, and risk estimation. However, the emphasized complexity, along with highly nonlinear behavior, dimensionality, and stochasticity, frequently lead to analytic intractability. Besides, traditional analytical models cannot replicate the actual behavior of a logistics system or supply chain. On the other hand, simulation models are not constrained by such restrictions and allow one to describe a logistics system with all its details and incorporate nonlinearity, uncertainty, and complexity. For instance, such traits as demand variability, lead-time fluctuations, multi-echelon inventory policies, and rare event occurrences can be considered within a simulation model \citep{ivanov_revealing_2018}.  Simulation plays the role of a dynamic, relatively low-cost testbed for conducting numerical experiments and scenario analysis. Such a testbed environment allows one to test strategies and scenarios that will lead to disruption or even push a system into extreme conditions revealing more about its structure, potential bottlenecks, and limits of resilience \citep{laguna_business_2018}.

Since early 2020, supply chains and production systems worldwide have experienced unprecedented shocks caused by the COVID-19 pandemic \citep{ivanov2023toward}. The pandemic catalyzed necessary transformations in the way supply chains are perceived and managed. This crisis exposed the vulnerabilities of traditional supply chains and prompted a global reassessment of how goods and services are produced, distributed, and consumed \citep{ivanov2023post}. As the world continues to recover and adapt, it is essential to rethink the status quo and approach supply chains as complex, intertwined supply networks that are more resilient to future disruptions. The post-pandemic adaptation necessitates a shift in perspective, moving away from a simple linear view of supply chains and embracing the intricacies of interconnected supply networks \citep{ivanov2020viability}. This change in mindset allows businesses and organizations to recognize the interdependencies among suppliers, manufacturers, and customers, as well as the importance of collaboration, transparency, and information sharing. By fostering stronger connections and building more robust networks, companies can better withstand unforeseen challenges and promote sustainable growth in a rapidly evolving global landscape \citep{ivanov2022industry}.

This flow of thought gave rise to a new concept of viability in supply chains \citep{ivanov2022viable}, which led to the development of an associated framework for assessing and enhancing supply chain resilience. This framework emphasizes the importance of adaptability, agility, and flexibility in supply networks, enabling businesses to respond effectively to disruptions and capitalize on new opportunities. By leveraging advanced technologies such as artificial intelligence, big data analytics, and the Internet of Things, companies can gain valuable insights, optimize their operations, and collaborate more effectively with partners throughout the supply network \citep{ivanov2022industry}. Embracing the concept of complex, intertwined supply networks and adopting the associated viability framework can help businesses navigate the post-pandemic landscape and build more resilient, adaptable, and sustainable supply chains for the future. One critical aspect of the viable supply chain concept is the use of scenario analysis, which plays a core role in designing flexible and resilient systems \citep{ivanov2020viability}. Rather than relying solely on a straightforward disruption magnitude analysis, this method emphasizes the importance of understanding the trajectories of how different disruption scenarios influence the severity of network degradation and recovery. By considering a wide range of potential outcomes, supply chain managers can identify vulnerabilities, implement mitigation strategies, and create more adaptive networks that are better equipped to handle future challenges.

Simulation modeling is a core technology behind scenario analysis and has become indispensable in designing and optimizing supply chains \cite{ivanov2019digital}. By employing simulation-driven decision-support systems, businesses can create proactive and resilient supply chain simulations that model dynamic behavior in the event of possible disruptions or disruption scenarios. This capability is essential in developing a supply chain that can achieve planned or actual acceptable performance even in the face of severe disruptions. The ability to analyze and understand complex supply chain systems enables stakeholders to identify weak points in their systems, implement strategies to mitigate potential risks, and create a more resilient and efficient supply chain overall. One of the major advantages of simulation models is their ability to model both the impact of disruptions on supply chain performance and the effectiveness of proactive resilience measures and recovery contingency plans \citep{ivanov2020coordination}. This comprehensive approach allows stakeholders to consider multiple factors that may affect their supply chain performance, such as changes in demand, inventory levels, transportation options, and supplier relationships. By examining these factors in tandem, businesses can gain a more holistic view of their supply chain operations and develop strategies that address the complexities of real-world environments.

Simulations also play an important role in testing optimal control algorithms, which can orchestrate supply chain operations in stochastic environments \citep{ivanov2022integrated, rolf2022review}. By experimenting with various control strategies in a virtual environment, businesses can fine-tune their processes and identify the most effective approach for their unique supply chain needs \citep{phadnis2022scenariocreation}. This not only helps to improve efficiency and reduce costs but also ensures that supply chains can adapt quickly and effectively to unforeseen events and uncertainties \citep{ivanov2022integrated}. The simulation-based methodology is particularly valuable in examining and predicting the impacts of "black swan" events, which are rare and unexpected occurrences with potentially severe consequences \citep{ivanov2020predicting}. By simulating these events, businesses can better understand their potential impacts and develop robust strategies to address them, thereby minimizing their negative effects on supply chain performance. This proactive approach to risk management allows businesses to anticipate and plan for a wide range of scenarios, ensuring that their supply chains remain resilient in the face of unexpected challenges \citep{phadnis2022scenario}. Additionally, simulation is a core component of Digital Twins, which are virtual replicas of physical systems or processes that allow for real-time monitoring and analysis \citep{saenz2022digital}. By integrating simulation models with Digital Twins, businesses can gain valuable insights into their supply chain operations, identify potential issues before they become critical, and rapidly test and implement changes to optimize performance. As a result, the combination of simulation modeling and Digital Twins enables a more comprehensive understanding of supply chain dynamics, empowering businesses to make informed decisions and drive continuous improvement in their operations \citep{ivanov2019digital}.

In short, simulation modeling serves as a powerful tool in the design and management of resilient supply chains, providing businesses with the necessary insights to navigate uncertain environments and mitigate the impact of disruptions. By leveraging advanced simulation techniques and integrating them with cutting-edge technologies such as Digital Twins, businesses can optimize their supply chain performance and ensure their ability to withstand even the most challenging circumstances. However, simulation modeling has several disadvantages. First, the development process is technically challenging, which implies that the domain experts cannot entirely develop such models, and the participation of simulation engineers or other technical specialists is required \citep{law_simulation_2000}. Second, technical specialists usually lack domain expertise or knowledge on the specific system or operation under consideration. This fact implies that the domain experts must first explain the real-world system and relations between all the important objects within the system \citep{leemis_discrete-event_2006}. The explanation, as a rule, includes the detailed verbal description and the conceptual model in the form of a flowchart or dependency graph. After that, the labor-intensive development process begins. Technical specialists have to transfer the verbal description and conceptual model into algorithmic implementation in the form of an executable computer program \citep{zeigler_theory_2019}. The described procedure assumes the back-and-forth information exchange between the domain experts and simulation engineers, which significantly increases the project’s duration and budget.

Considering this problem, our study explores how the latest advancements in Artificial Intelligence (AI) and Natural Language Processing (NLP) can be applied to develop fully-functional and executable simulation models of the systems vital to supply chains. At this point, it is essential to define terms related to AI and NLP crucial for further reading. These definitions are preliminary and will be further formally unfolded in the background and methodology chapter. In this study, we take an instrumentalist view of AI \citep{bostrom_superintelligence_2014}, and do not speculate on AI cognition or the philosophical aspects. The definition of AI proposed by Arthur \cite{samuel_studies_1959} suits well in the context of our study. \emph{“AI is the field of study that gives computers the ability to learn by using sampled data without being explicitly programmed.”}. NLP, in its turn, is a branch of AI that encompasses various language-related tasks, including sentiment analysis, speech recognition, machine translation, text generation, and summarization \citep{li_language_2022}.

\begin{figure}[!htb]
  \centering
  \includegraphics[scale=0.58]{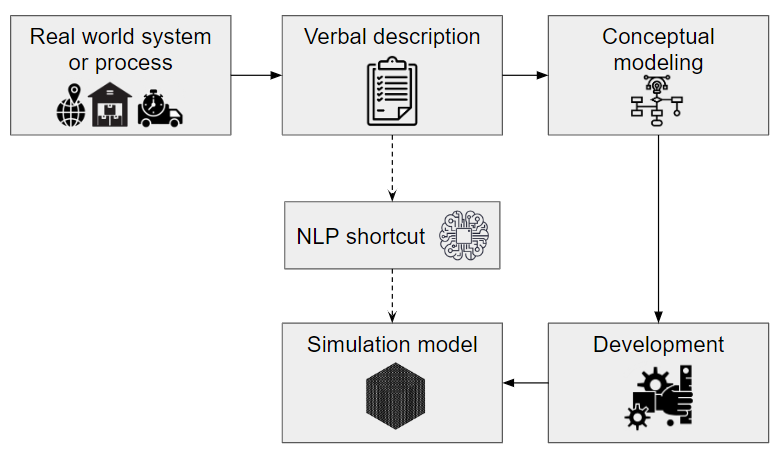}
  \caption{“NLP shortcut” means that such labor-consuming steps as conceptual modeling and coding can be potentially skipped by taking advantage of the state-of-the-art NLP techniques.}
  \label{fig:fig1}
\end{figure}

Our study focuses on the proof of concept that the “NLP shortcut” exists in principle, and valid simulation models of relatively simple queuing and inventory control systems can be produced automatically based on the verbal description (See Figure \ref{fig:fig1}). Therefore, we postulate the following research question: “Can simulation models of logistics systems be produced automatically given the verbal description in natural language?”.

The remainder of this paper is organized as follows. Section \ref{related_work} reviews some works related to automated code generation and programming using GPT-3 and expert systems based on NLP. Section \ref{background} sheds light on background and methodology, focusing on the state-of-the-art Transformer-based language models. Experimental results and analyses are presented in Section \ref{experiments}. Section \ref{discussion} presents the discussion regarding the GPT-3 Codex capabilities, ongoing trends in NLP, and their potential implications for developing simulation models. Finally, our contribution, as well as promising directions for future research, are summarized in Section \ref{conclusions}.

\section{Related Work}
\label{related_work}
Our work is related to two streams of research: automated code generation and programming using GPT-3, and decision-support and expert systems based on NLP. The background and methodology chapter covers language modeling in general and state-of-the-art Transformer-based models.

\subsection{Automated code generation using GPT-3}

The topic of automated code generation using language models has recently attracted research attention \citep{clement_pymt5_2020}. Since the GPT-3 Codex private beta program was launched on August 10, 2021, the first attempts of the academic community to explore the language model’s capabilities have been published as preprints. 

Notable examples include \cite{hocky_natural_2021}, who explored the capabilities of the GPT-3 Codex to solve simple computational chemistry problems. \cite{drori_solving_2021} have applied the GPT-3 Codex to solve MIT’s Linear Algebra course and Columbia University’s Computational Linear Algebra course with perfect accuracy. \cite{shporer_learning_2022} managed to combine the GPT-3 Codex with graph neural networks to solve computational problems from the Introduction to Astronomy course. \cite{trummer_codexdb_2022} was able to finetune GPT-3 Codex to translate user-provided instructions and descriptions of database properties into SQL queries.

Summarizing these first attempts, we can observe a general tendency for domain scientists to propose ways to adapt and apply GPT-3 Codex to their field of expertise. Since there are no applications in logistics yet, we continue this tendency by applying GPT-3 Codex to build simulation models of logistics systems. Summarizing these works, we can observe that expert and decision support systems take advantage of the state-of-the-art NLP algorithms to apply such downstream NLP tasks as semantic search, natural language generation, text classification, and sentiment analysis to domain-specific problems.

\subsection{NLP-based systems}
\begin{figure}[!htbb]
  \centering
  \includegraphics[scale=0.55]{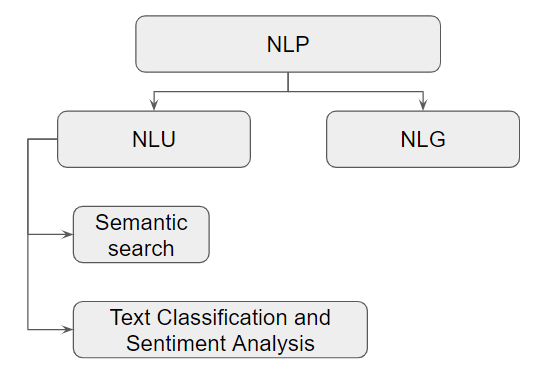}
  \caption{Classification scheme of the relevant publications.}
  \label{fig:Class}
\end{figure}

Another stream of related research incorporated decision-support and expert systems based on NLP. Because of rapid digitalization and unprecedented opportunities for harvesting and storing textual data NLP has recently experienced substantial breakthroughs in machine translation, pattern matching, sentiment analysis, and speech recognition. These breakthroughs have not only improved people’s daily lives but also notably revolutionized decision-support and expert systems \citep{kang_natural_2020}. Figure \ref{fig:Class} provides a classification scheme for the most relevant publications in the context of our study. A recent systematic review sheds light on the diversity of NLP applications in management and business \citep{kang_natural_2020}.

\FloatBarrier

NLP includes two primary research directions: Natural Language Understanding (NLU) and Natural Language Generation (NLG). The primary role of NLU is to comprehend the natural language by mining text documents and extracting valuable information for downstream tasks \citep{schank_conceptual_1972}. In contrast, NLG focuses on the production of text in natural language, which is comprehensive for humans \citep{reiter_building_1997}. It is essential to highlight that GPT-3, along with the other state-of-the-art language models, are capable of both.

\subsubsection{Semantic search}
Semantic search is a searching technique in which a search query aims to not only find and extract relevant keywords but to determine the searcher's intent and the contextual meaning of the words \citep{guha_semantic_2003}. 

Notable examples include \cite{habernal_swsnl_2013}, who presented the NLP system for a semantic web search. The developed system is end-to-end in the sense that it includes preprocessing, semantic analysis, semantic interpretation, and executing a sequence of queries to retrieve the results. The system demonstrated solid NLU capabilities with written texts in Czech and English across the domain of public transportation. \cite{sangers_semantic_2013} proposed an NLP-based search engine for finding semantic Web services. The framework based on this search engine was capable of matching relevant keywords gathered from Web service descriptions and the user query. \cite{lupiani-ruiz_financial_2011} presented a semantic search engine for financial news. The search engine was combined with an ontology population tool to assist in keeping the financial ontology up-to-date. Besides, the expert system was equipped with a module for crawling the Web and annotating financial news according to the financial ontology.

\subsubsection{Text classification and sentiment analysis}
Text classification is the NLP task of assigning a label or class to a given fragment of text. The most common use case of text classification includes sentiment analysis, an NLP technique that focuses on the analysis of people’s opinions, sentiments, appraisals, attitudes, and intents expressed in written text. Sentiment analysis is increasingly important in business and society and is often performed on textual data to assist businesses in monitoring product sentiment and understanding customer opinion \citep{liu_sentiment_2012}.

Notable examples include \cite{prollochs_negation_2016}, who designed a decision support system equipped with NLP for financial news sentiment analysis. The system incorporated reinforcement learning, hidden Markov models, conditional random fields, and rule-based methods and could achieve a substantial improvement in the correlation between sentiment and stock return. \cite{zinovyeva_antisocial_2020} proposed an expert system to automate content monitoring and antisocial behavior detection. The system incorporated advanced NLP techniques, including bidirectional encoding, attention mechanisms, hierarchical text representations, and Transformer-based language models. \cite{abbasi-moud_tourism_2021} developed a tourism recommendation system capable of extracting users' preferences to provide personalized recommendations. The system analyses semantically clustered and sentimentally analyzed user reviews to detect a tourist's preferences and suggest the most matching options.

\cite{liu_large-scale_2019}combined such NLP techniques as Word2Vec and n-gram to track individual-level purchasing behaviors on an e-commerce site to classify customers based on their expectations for recovery service. \cite{li_knowledge-oriented_2019} proposed a knowledge-oriented convolutional neural network to extract and classify causal relations of the form product-producer, employer-employee, cause-effect, and message-topic. The proposed expert system includes both humans to capture the linguistic clues of causal relationships and a data-oriented component capable of learning other important features from the data. \cite{kim_transparency_2020} proposed a framework that adopts both NLP and computer vision for explaining and visualizing convolutional neural networks for text classification. 

\subsubsection{Natural language generation}
NLG can be defined as the subfield of NLP concerned with constructing computer systems capable of producing understandable texts in English or other languages from some underlying representation of information.
Notable examples include \cite{wulf_natural_2017}, who incorporated NLG techniques into a decision support system to enhance the understanding and traceability of multi-criteria decision-making. The study concludes with the statement that the NLG approach is especially beneficial for complex interpretational tasks. \cite{lee_natural_2018} proposed the deep learning-based encoder-decoder model for clinical decision support. The model was capable of generating synthetic electronic health records that appear to preserve accurate epidemiological information. \cite{garcia-mendez_library_2019}  developed an NLG system for the automatic generation of administrative reports. The system could generate complete, coherent, and correctly spelled sentences.

\subsubsection{Conclusion}
Summarizing these works, we can observe that expert and decision support systems take advantage of the state-of-the-art NLP algorithms to apply such downstream NLP tasks as semantic search, natural language generation, text classification, and sentiment analysis to domain-specific problems.

\section{Background and Methodology}
\label{background}
The best technical approaches of the recent past assumed that NLP systems should specialized on a narrow range of tasks \citep{cambria_jumping_2014}. However, the rules of the game changed with the advent of new deep learning architecture, the Transformer \citep{vaswani_attention_2017}. GPT-3 is a notable example of Transformer-based NLP systems and the main focus of our study. 

\subsection{Language Model}
Language models are at the core of contemporary NLP methodologies. Language modeling can be postulated in the purely unsupervised learning setting \citep{radford_language_2018}. At first glance, such an assumption appears to be too strong. Especially considering the richness, creativity, and elegancy of natural language pieces produced by such masters as William Shakespeare, Barbara Cartland, Brothers Grimm, and Leo Tolstoy. Fortunately, due to the tempo, and exigencies, the vast majority of human utterances are significantly simpler, repetitive, frequently self-similar, and, as a result, far more predictable. Programming languages are even more regular and structured. For example, \cite{hindle_naturalness_2016} demonstrated that repetitive patterns occur at lexical, syntactic, and semantic levels in code corpora. 

At this point, a definition of the term corpora is essential, given that it is central for NLP and linguistics in general. The corpora can be defined as a collection of linguistic data consisting of structured, processed, and digitally stored texts \citep{rocha_sentence_2014}. Formally speaking, the goal of learning is reduced to the estimation of distribution from corpora given a set of training examples \((x_1, x_2, ..., x_n)\), such that every example is a sequence of symbols of variable length \((s_1, s_2, ..., s_n)\). Since both natural and programming languages are ordered sequentially, the joint probabilities over symbols (or tokens) can be factorized as the product of conditional probabilities:
\begin{equation} \label{eq:1}
p(x) = \prod_{i=1}^{n} p(s_n | s_1, s_2, ..., s_{n-1})
\end{equation}
This assumption allows one to perform tractable sampling and estimation of \(p(x)\) and other conditional probabilities of the form \(p(s_{n-k},..., s_n|s_1,..., s_{n-k-1})\) \citep{bengio_neural_2003}.

To conclude, language models are statistical models that estimate the probability of a given sequence of tokens, for example, words or code-specific syntax, to appear in the text with respect to the context. 

\subsection{Transformer}
\begin{figure}[!htb]
  \centering
  \includegraphics[scale=0.42]{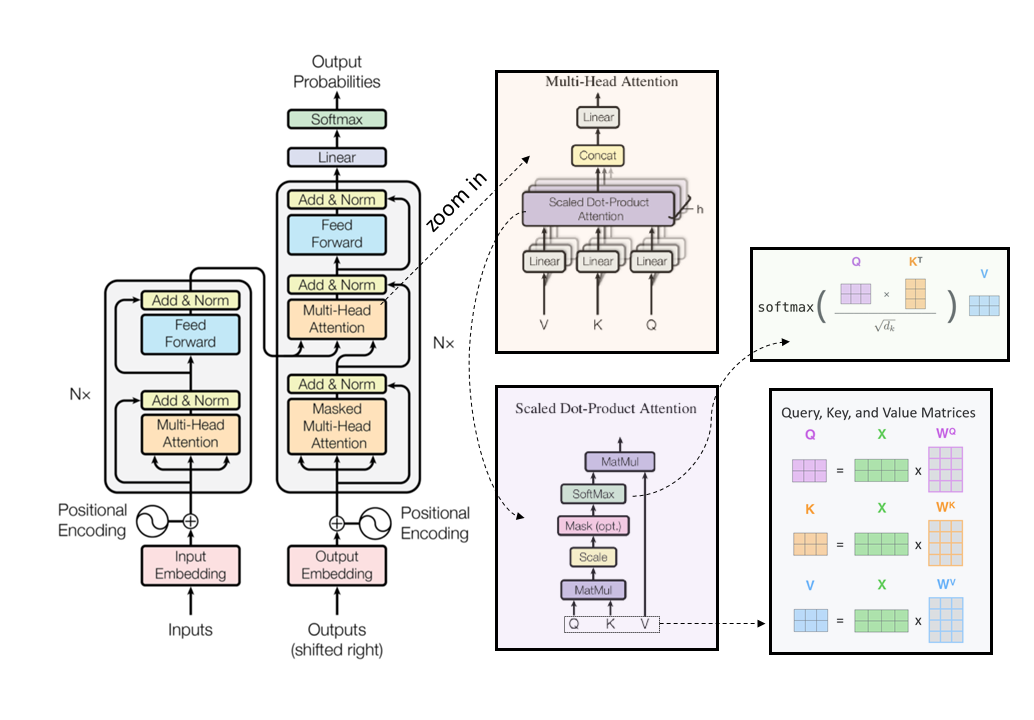}
  \caption{The Transformer’s architecture. Multi-Head Attention consists of several attention layers running in parallel. Query, Key, and Value matrices are calculated by packing the embeddings into a matrix \(X\), and multiplying it by the weight matrices \(WQ\), \(WK\), \(WV\). The illustration is based and adapted from \cite{vaswani_attention_2017} and \cite{alammar_illustrated_2018}.}
  \label{fig:fig3}
\end{figure}

In the recent five years, there have been significant improvements in the performance of language models based on conditional probabilities of the form (See Equation \ref{eq:1}). The most notable advances can be attributed to self-attention architectures like the Transformer \citep{vaswani_attention_2017}. 
Encoder-decoder structure constitutes a core behind the Transformer’s architecture. The role of the encoder is to map an input sequence of symbol representations \((x_1, x_2, ..., x_n)\) to a sequence of representations \((z_1, z_2, ..., z_n)\). The decoder produces an output \((y_1, y_2, ..., y_n)\) given \((z_1, z_2, ..., z_n)\) as an input. This process is autoregressive in the sense that the previously generated symbols are used as additional input \citep{graves_generating_2014}. Both encoder and decoder use stacked self-attention mechanisms as well as fully connected layers (See Figure \ref{fig:fig3}). 
\FloatBarrier

The encoder consists of a stack of \(N\) identical layers. Each layer includes two sub-layers. The first sublayer is a multi-head self-attention mechanism, and the second one is a position-wise fully connected feed-forward network akin to multilayer perceptron architectures. All the sub-layers and embedding layers have an output of the same dimension \(d_{model}\) in order to facilitate the residual connections. The decoder is structured in a similar way including a stack of \(N\) identical layers. 
A self-attention function is intended to map a query and key-value pairs to an output, where the query, keys, values, and output are vectors. The output is computed as a weighted sum, where the weights associated with the corresponding values are computed using a feed-forward artificial neural network. 
Scaled Dot-Product Attention is a fundamental component in the self-attention mechanism. For the sake of computational efficiency, the attention function is computed on a set of queries in parallel. The input consisting of queries and keys of dimension \(d_k\), and values of dimension \(d_v\) are stacked into matrices \(Q\), \(K\), and \(V\).
\begin{equation} \label{eq:2}
Attention(Q, K, V) = softmax(\frac{QK^T}{\sqrt{d_k}})
\end{equation}
The \(Q\), \(K\), and \(V\) matrices are linearly projected \(h\) times to \(d_k\), and \(d_v\) dimensions, respectively. After that, the self-attention function is performed on each of these projections, resulting in \(d_v\)-dimensional output vector. As a result, multi-head attention allows the model to access information from different representation subspaces at different positions. 
\begin{equation} \label{eq:3}
MultiHead(Q, K, V) = Concat(head_1,..., head_h)W^O
\end{equation}
where \(head_i = Attention(QW_i^Q, KW_i^K, VW_i^V)\) with parameter matrices \(W_i^Q \in \mathbb{R}^{d_{model} \times{d_k}}\), \(W_i^K \in \mathbb{R}^{d_{model} \times{d_k}}\), \(W_i^V \in \mathbb{R}^{d_{model} \times{d_v}}\) and \(W^O \in \mathbb{R}^{hd_{v} \times{d_{model}}} \) \citep{vaswani_attention_2017}.
Besides the sub-layers, each of the layers in both encoder and decoder contains a fully connected feed-forward network that can be represented as a composite function \(FFN(.)\). \(FFN(.)\) is equipped with Rectified Linear Unit (ReLU) activation function and includes two linear transformations.
\begin{equation} \label{eq:4}
FFN(x) = max(0, xW_1 + b_1)W_2+ b_2
\end{equation}
The Transformer takes advantage of the learned embeddings to convert the input and output tokens to vectors of the dimension \(d_{model}\), which is basically the embedding size. Besides, the regular learned linear transformation and \(softmax(.)\) activation function convert the decoder output to estimated probabilities of the next token to appear in the sequence.

\subsection{GPT Family and Codex}
\begin{figure}[!htb]
  \centering
  \includegraphics[scale=0.17]{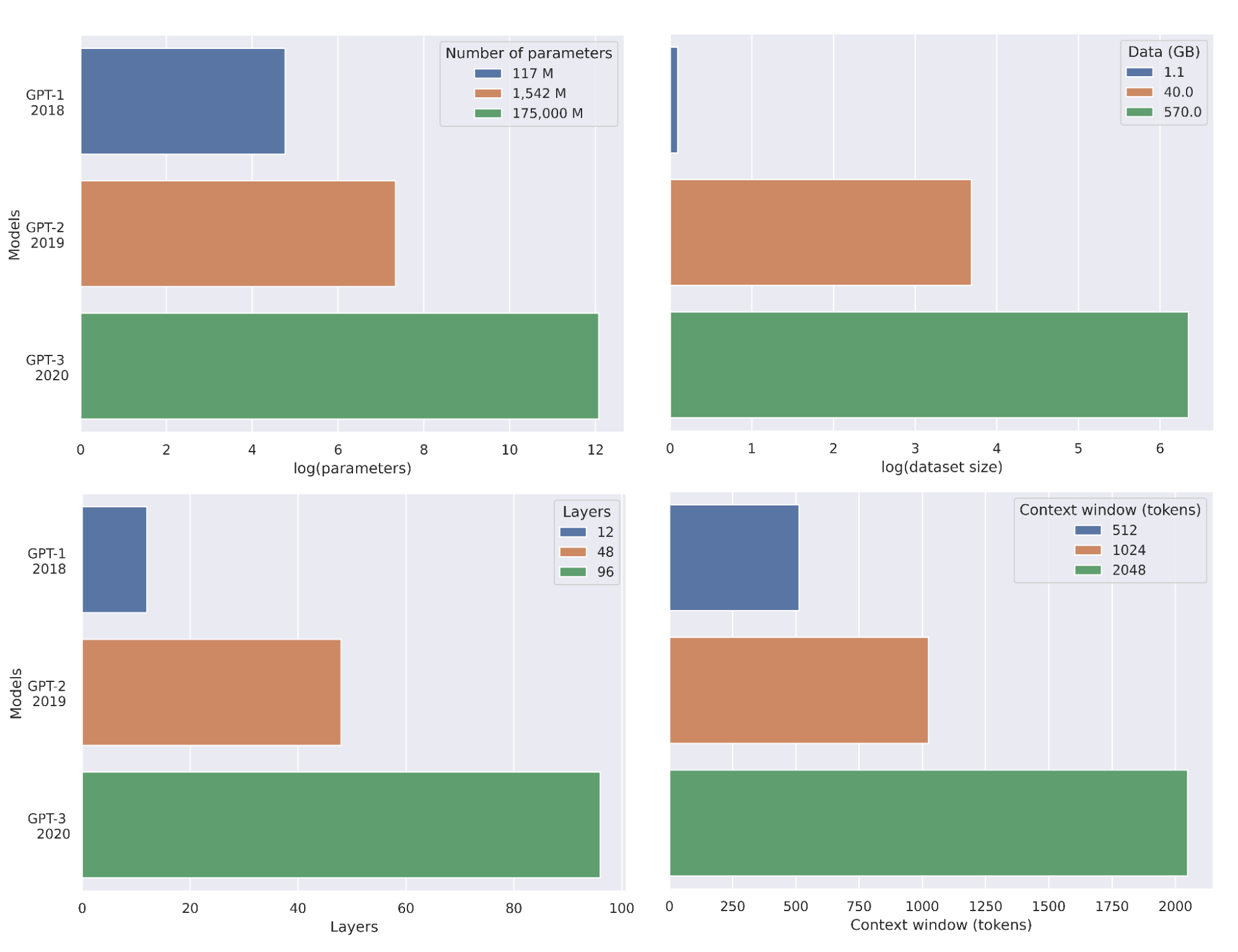}
  \caption{The size of the GPT-3 models increases rapidly in terms of the number of layers, parameters, training data volumes, and the width of the context window.}
  \label{fig:fig6}
\end{figure}

Shortly after the advent of the transformer language model, such a mega-scale statistical brute-force approach was widely adopted by the NLP community. As a result, the scale of language models has increased substantially (See Figure \ref{fig:fig6}). From 100 million parameters and 1.1 GB of training data in GPT-1 to 1.5 billion parameters and 40 GB in GPT-2 \citep{radford_language_2018}, and finally 175 billion parameters and over 570 GB in GPT-3 \citep{brown_language_2020}.
\FloatBarrier

Given such numbers and data volumes, the prominent question arises, “Does the size of the language model affects its performance?” and the short answer is “Yes!”. \cite{brown_language_2020} conclude with the statement that increases in model parameters entail improvements in text synthesis. During unsupervised pretraining on the immense text corpora, language models acquire a plethora of non-trivial pattern recognition abilities and, as a result, develop an incredibly diverse skill set, including reading comprehension, textual entailment, question answering, grammar correction, and text auto-completion. 

The rapidly increasing scale of language modeling has significantly fueled the interest in the longstanding challenge of automated computer code generation and program synthesis. Even the preliminary experiments with GPT-3 revealed its ability to produce simple executable computer programs in Python \citep{brown_language_2020}. After the initial success of such language models as CodeBERT \citep{feng_codebert_2020} and PyMT5 \citep{clement_pymt5_2020}, the OpenAI team hypothesized that it is possible to fine-tune GPT on code from GitHub to produce functionally correct code bodies given natural language document strings \citep{chen_evaluating_2021}.
The OpenAI team has fine-tuned several GPT models on code. The largest one, Codex, contained over 12 billion parameters and the doubled token window of 4,096 tokens \citep{zaremba_openai_2021}. The training data was collected from 54 million public GitHub repositories, containing 179 GB of unique Python files under 1 MB each.  In order to test the resulting model, a new HumanEval dataset of 164 hand-written programming problems has been proposed. As a result, Codex has displayed non-trivial performance and the ability to solve the majority of the problems from the HumanEval dataset.

\section{Experiments}
\label{experiments}
This chapter is structured as follows. First, the OpenAI API is introduced. Second, the framework based on the OpenAI API is proposed. Last, the framework is applied to generate simulation models of inventory and queuing systems given the description in English. English is chosen because GPT-3 is the most proficient in it \citep{brown_language_2020}. Besides, English is the international language of engineering and science that most potential users should be able to understand. Python is selected for the same reason, GPT-3 Codex is the most proficient in Python 3 \citep{chen_evaluating_2021}.

\subsection{OpenAI API}
\begin{figure}[!htb]
  \centering
  \includegraphics[scale=0.4]{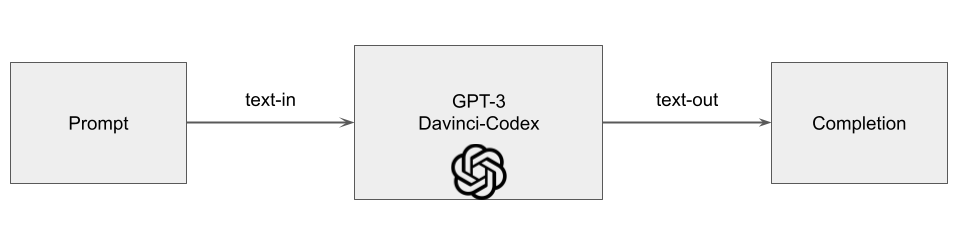}
  \caption{OpenAI API provides the straightforward text-in-text-out interface for the wide range of transformer-based language models.}
  \label{fig:fig7}
\end{figure}

The OpenAI API offers a broad spectrum of transformer-based language models (also known as engines). The framework proposed in our paper takes advantage of the Davinci-codex engine since it has the largest token window of 4,096 tokens and is the most capable of translating natural language to code and back \citep{brockman_openai_2021}.
The OpenAI API is a flexible interface that uses human-readable text as both input and output. The input text is referred to as a prompt, and the returned text is called completion. A prompt is used by GPT-3 models as the baseline to determine the task and generate a text completion in the attempt to match the underlying structure, logic, context, or pattern (See Figure \ref{fig:fig7}).
Prompts are akin to programming using English in the sense that a user has to know exactly what he or she is trying to accomplish, but rather than writing code, one can use words and plain text. Technically, any text can be used as a prompt. However, in order to achieve the desired outcome and unlock GPT-3's true potential, the prompts have to be engineered effectively. 

\begin{figure}[!htb]
  \centering
  \includegraphics[scale=0.45]{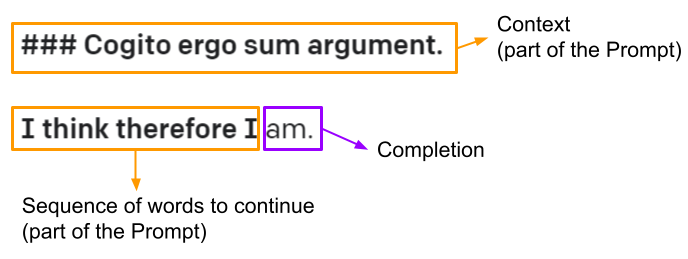}
  \caption{The language model finishes the sentence given the beginning of the sentence and context as the prompt.}
  \label{fig:fig8}
\end{figure}

Prompt engineering is a very young field of study. However, several well-established practices have already formed \citep{zhao_calibrate_2021, reynolds_prompt_2021, hocky_natural_2021}. One of them, applied in our study, is to separate the context, instructions (specific task to perform), and pattern to continue. In the most general sense, the pattern can be defined as the sequence of words to continue. Figure \ref{fig:fig8} shows how GPT-3 finishes the “first principle” of Rene Descartes's philosophy, namely “I think, therefore I am.” given the beginning of the sentence and context as the prompt \citep{descartes_descartes_1990}. 

\begin{figure}[!htb]
  \centering
  \includegraphics[scale=0.5]{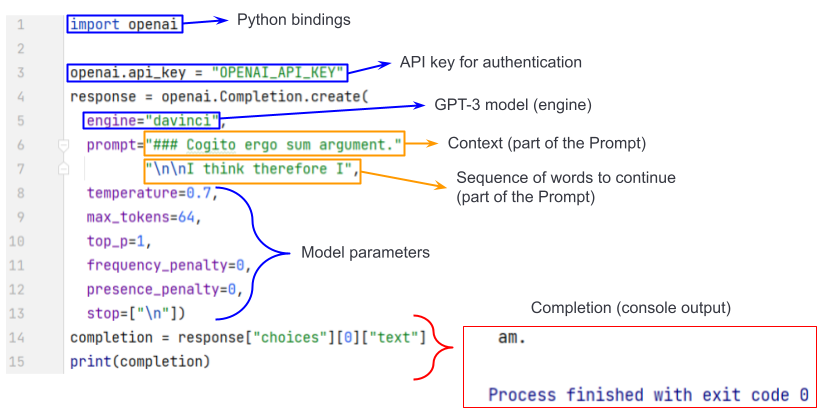}
  \caption{Pythonic interface to OpenAI API.}
  \label{fig:fig9}
\end{figure}

OpenAI API is accessible using Python bindings, which is a convenient interface for conducting experiments and building on top (See Figure \ref{fig:fig9}).
\FloatBarrier

\subsection{Experiment design}
The completion quality depends significantly on the prompt structure, context specification, and explicit task description. The well-engineered prompt helps to improve the language model predictions and unlock GPT-3's true potential. Prompt engineering at the current stage is mainly performed through trial and error. However, several core principles can significantly accelerate the process. First, it is helpful to try different formulations of the same prompt, which might appear similar to humans, but not for the language model. Such a phenomenon arises because the model could learn that the different formulations are, in fact used in different contexts and for different purposes. Additionally, it is crucial to provide the model with sufficient context. Besides prompt engineering, hyperparameter tuning is essential as in any other state-of-the-art machine learning technique.

\begin{figure}[!htb]
  \centering
  \includegraphics[scale=0.45]{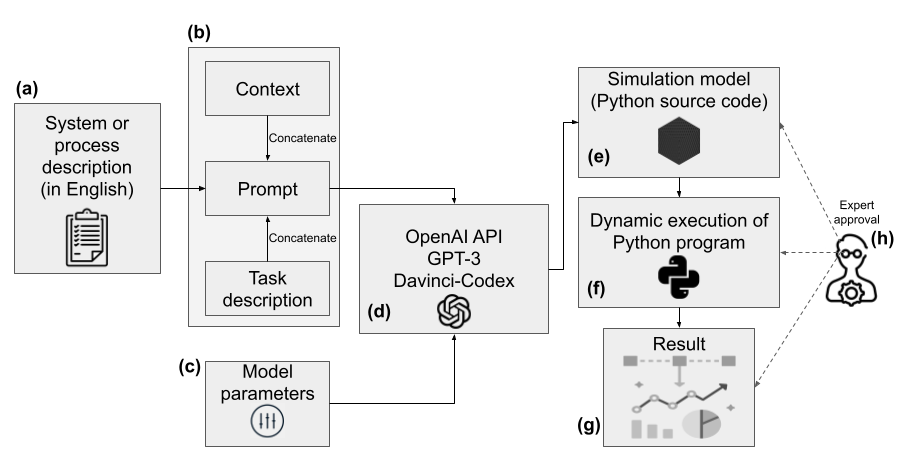}
  \caption{The proposed framework generates  and executes the simulation model based on its description in English.}
  \label{fig:fig16}
\end{figure}

In order to semi-automate the experiments we have implemented a high-level framework on top of OpenAI API. The framework is designed to answer the original research question and verify whether the language model is capable of translating a verbal description of the system to executable simulation in Python (See Figure \ref{fig:fig16}). 
\FloatBarrier

The description of the inventory and queuing systems in English is used as an input to which context and task description are appended using the concatenation operator (a). The language model uses the intermediate result as a prompt (b). After that, the parameters of the language model are adjusted in such a way (c) that the language model (d) produces executable Python code that corresponds to the simulation model of the specified logistics system (e). Python is chosen for two main reasons. Firstly, according to \cite{chen_evaluating_2021}  GPT-3 Codex is the most proficient in Python 3. Secondly, the Python programming language is distinguished by the dynamic execution (f). Namely, the \emph{“exec”} and \emph{“eval”} methods can execute code obtained during a program’s run. The proposed framework uses the \emph{“eval”} method that parses the string, evaluates it as a Python expression, and compiles it into bytecode. Bytecode is a low-level platform-independent set of instructions for the Python Virtual Machine, where the expression is run, and the value of the expression is finally returned (g). After that, the validity of the generated source code and the generated output is verified by a human expert (h). There are two major ways how this principle can be implemented in expert systems. First, such pivot steps as code generation, execution, and production of the visual output can be performed within a single runtime and verified by an expert after that. Second, they can be implemented sequentially such that the next step is performed only after the expert’s approval. The following experiments have been conducted following the first way. The second way is more suitable for “human-in-the-loop” style expert systems.

Since the general capabilities to generate valid Python code were demonstrated by the original OpenAI paper \citep{chen_evaluating_2021}, our study focused exclusively on simulation models of inventory and queuing systems written in Python. The GPT-based framework has been finetuned in order to produce valid executable Python code based on the descriptions of a single-server queuing system and single-product inventory-control system. It is essential to highlight that unlike in classic NLP problems, such metrics as perplexity and BLEU score \citep{papineni_bleu_2001} can not evaluate the validity of code generation and program synthesis. Besides, it is impossible to rely on automated testing since simulation models under consideration are dynamic, iterative systems. Therefore, the logic behind the code structure is as important as the simulation output. In this regard, all the test problems have been verified manually by the authors of this paper. 

\subsection{Results}
The experiments were conducted in a reproducible manner and the source code is available in the GitHub repository\footnote{\url{https://github.com/Jackil1993/GPT3_SCM}}.

During the experiments with prompt engineering and hyperparameters tuning two efficient approaches were discovered. According to the first approach, the language model produces a simulation of a single-product inventory-control system given the domain context, detailed description of the process, and a list of variables with the corresponding values. 

\begin{figure}[!htb]
  \centering
  \includegraphics[scale=0.32]{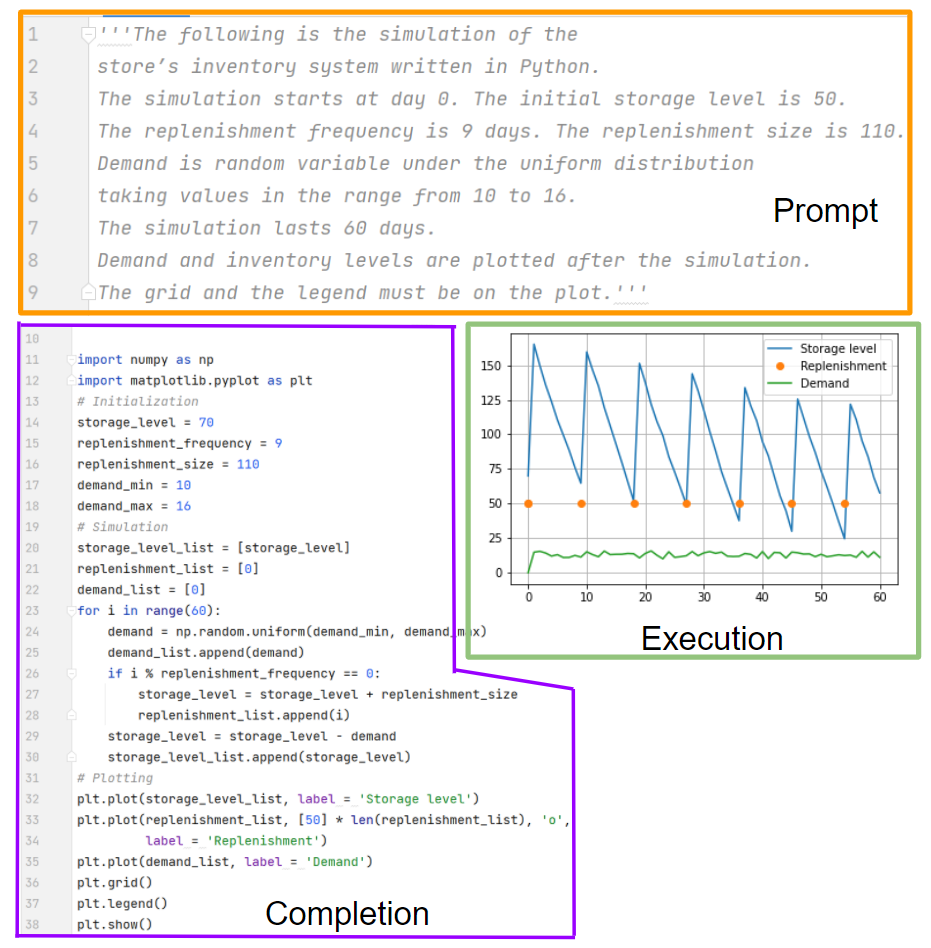}
  \caption{Given the set of instructions in English, GPT-3 Codex imports such Python libraries for random number generation and visualization, declares variables, implements the simulation logic inside the for loop, and plots the inventory dynamics at the end.}
  \label{fig:fig17}
\end{figure}

\begin{figure}[!htb]
  \centering
  \includegraphics[scale=0.32]{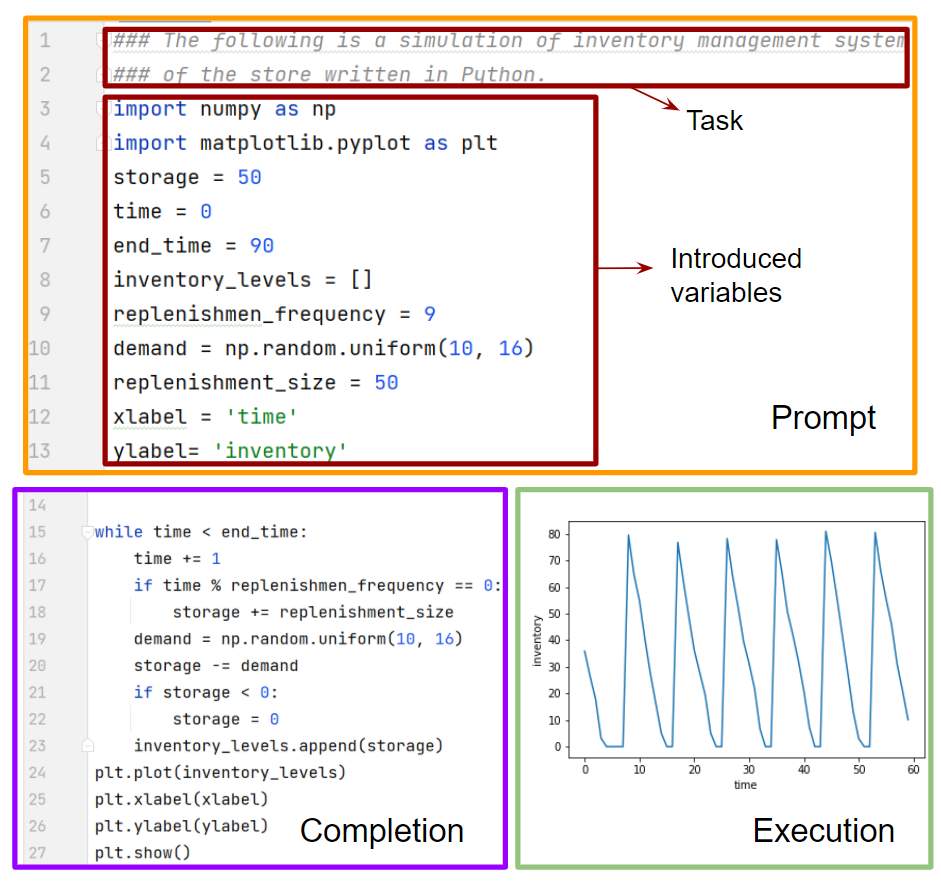}
  \caption{Given the one-sentence specification of the inventory management context and the list of variables and data structures to utilize in the code, GPT-3 Codex can implement the simulation correctly.}
  \label{fig:fig18}
\end{figure}

Figure \ref{fig:fig17} shows that the mentioned instructions are enough to produce a valid executable simulation model and visualize the inventory dynamics. GPT-3 Codex starts with importing Python libraries for random number generation to model stochasticity and data visualization to visualize inventory dynamics at the end. After that, all the necessary variables and data structures are defined. It is worth mentioning that GPT-3 Codex names variables in a meaningful way, preserving the domain-specific context and keeping the code human-readable. After that, the logic behind inventory control is correctly implemented. First, random demand under uniform distribution is generated, simulating the market uncertainty. Second, the replenishment of the previously ordered batch is performed, increasing the current inventory level by the order quantity. Third, demand is satisfied if there is sufficient inventory on hand. These core steps are executed in the loop such that each iteration corresponds to a day. The algorithm keeps track of the variables’ changes in order to plot the dynamics at the end. The resulting figure is human-readable and contains both grid and legend. It is worth mentioning that GPT-3 Codex decided to display the replenishment days on the figure, which had not been mentioned in the original task but was quite helpful from a data-visualization point of view. This decision could be explained by the fact that GPT-3 Codex was trained on a substantial part of GitHub that contained code samples of simulation models of similar inventory control systems. Therefore, the language model understands that human experts tend to visualize replenishment days when developing such simulations.
\FloatBarrier
However, there is an alternative approach. Instead of a detailed task description, the language model can use the short one-sentence specification of the inventory management context and the list of variables and data structures to utilize in the code. Since variables are named in a meaningful way with regard to the domain context, GPT-3 Codex can use them in the code correctly. For example, given such variables as \emph{xlabel=’time’} and \emph{ylabel=’inventory’} GPT-3 Codex decides to plot inventory levels against time at the end of a simulation run without being explicitly asked to do so. That is where the true power of proficiency in both Python and English is revealed. Unlike in the previous example, the language model implemented the simulation using a while loop, which is functionally equivalent. Besides, since GPT-3 was not asked to plot the dynamics in a specific way, the resulting figure lacks the grid and legend. Nevertheless, the figure is human-readable, and the dynamic pattern typical for inventory levels is recognizable (See Figure \ref{fig:fig18}). 

\begin{figure}[!htb]
  \centering
  \includegraphics[scale=0.32]{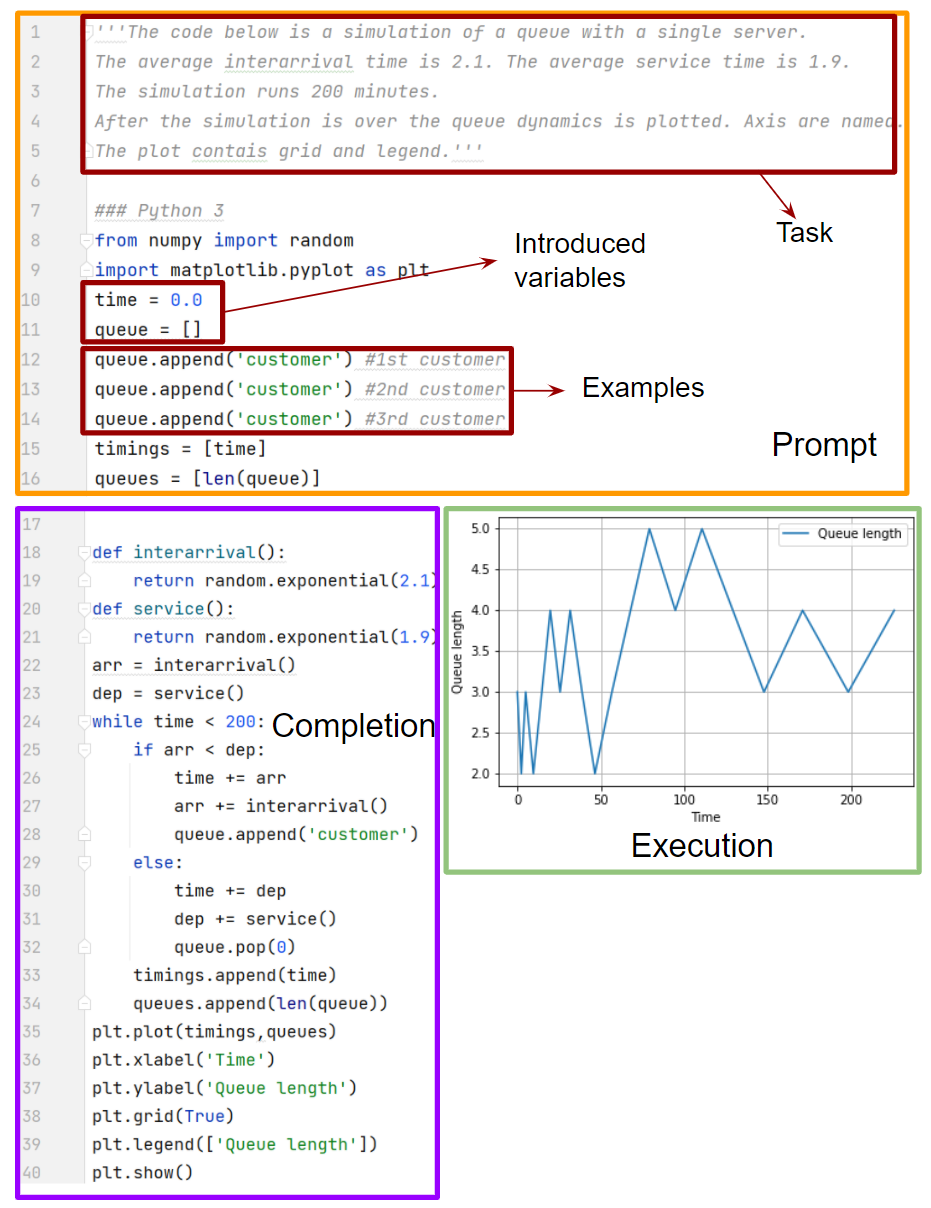}
  \caption{Given the detailed task description, variables to utilize, and a few examples, GPT-3 Codex could produce a valid simulation model of a single-server queueing system.}
  \label{fig:fig19}
\end{figure}

The discovery of the alternative approach was crucial for generating the prompt that entails the valid simulation of the single-server queuing system. The framework could not generate the correct simulation given only the description of the queuing system. As a result, the valid simulation model was obtained by combining the elements of both approaches. Namely, the language model received the detailed task description as well as the variables to utilize. Besides that, it was necessary to provide a model with an example on how to handle the arrival of customers to the system. Eventually, GPT-3 Codex could implement the simulation logic by executing the arrival and departure events within the while loop. The simulation schedules the timing of next arrival and departure, either adding or removing a customer from the system depending on which event happens first (See Figure \ref{fig:fig19}). This principle is well-known as Discrete Event Simulation \citep{law_simulation_2000}. This discovery could also be explained by the fact that GPT-3 Codex was trained on a substantial part of GitHub that contained code samples of various simulation models, and some of them were implemented according to the Discrete Event Simulation paradigm.
\FloatBarrier

Summarizing, we can come to a conclusion that GPT-3 Codex is proficient in Python as well as in English and as a result is capable of grasping both the core principles behind the simulations of logistic systems and the domain-specific context. Additionally, the framework built on top of the finetuned language model can provide the “NLP shortcut” for simulation modeling. Specifically, the framework can produce simple but functionally valid simulations of queuing and inventory control systems given the verbal description in English.

\section{Discussion}
\label{discussion}
GPT-3 Codex demonstrated convincing expertise in Python as well as an understanding of the logistics systems and domain-specific vocabulary.  Moreover, the language model could produce functionally correct simulations of queuing and inventory control systems given the verbal description in English.
At this point, the capabilities of the frameworks built on top of the language model are limited to simple toy-like problems. These limitations are subject to the number of model parameters, size and quality of training datasets, and the context window size. However, considering the rapid progress in the NLP field, significantly more complicated problems can be approached in the near future. It is important to point out that since prompt engineering plays the central role in the proposed framework, more powerful text-in-text-out language models can be effortlessly incorporated without changing the core principle. 

The recent history of NLP may shed the light on the progress in the field. Namely,  shortly after the advent of the transformer language model, such a mega-scale statistical brute-force approach based on transformer architectures was widely adopted by the NLP community. As a result, the scale of language models has increased substantially in the last few years, from 100 million parameters and 1.1 GB of training data in GPT-1 to 175 billion parameters and over 570 GB of data in GPT-3 \citep{brown_language_2020}. Despite our study using GPT-3 as the primary language model, while we were writing this manuscript, DeepMind developed a 280 billion parameter transformer language model called Gopher \citep{borgeaud_improving_2021} trained on the dataset that contains 2.35 billion documents, or about 10.5 TB of text \citep{rae_scaling_2021}. Besides, Google Brain presented the Switch Transformer architecture with almost 1.6 trillion (1571 billion) parameters trained on the 750 GB  dataset \citep{fedus_switch_2021}. Therefore, we expect that the size and complexity of simulations that the framework, designed according to the same principles as one proposed in this paper, can produce will notably increase over time. 

It is also important to mention the impressiveness of scaling and adoption rate. Namely, on March 25, 2021 (less than nine months since the launch of GPT-3) OpenAI team reported that the model generates more than 4.5 billion words per day. Additionally, the number of applications leveraging GPT-3 has exceeded 300 at that time \citep{zaremba_openai_2021}. Taking into consideration these tendencies, the futuristic mission of the OpenAI team to develop artificial general intelligence, \emph{“highly autonomous capable of outperforming humans at most economically valuable work,”} does not seem to be overambitious anymore \citep{openai_gpt-3_2018}. Keeping in mind the potential negative implications of automation in general and synthetic code generation in particular \citep{schwab_fourth_2016, acemoglu_automation_2019, smith_post-automation_2021}, we nevertheless wish to emphasize that simulation engineers do not spend all time writing code. Other essential duties include conferring with stakeholders, writing specifications, and documenting existing simulation models. The tools developed based on the proposed concept have the potential to remove the tedium of programming and let simulation engineers along with the domain scientists focus on the high-level consideration of the problem and holistic thinking. It is worth highlighting that the technological progress and better tools have not reduced the need for engineers and scientists over time but rather expanded the complexity, scale, and scope of problems that can be approached by a single expert or a research group.

\section{Conclusion, managerial implications, and future research}
\label{conclusions}
This section concludes the research, provides managerial implications, and outlines promising directions for future research. 

\subsection{Conclusion}
Our work is the first attempt to apply NLP to partially automate the development of simulation models in general and simulations of queuing and inventory control systems in particular. We demonstrated that GPT-3 Codex is proficient in Python as well as in English and, as a result, is capable of understanding both the core principles behind the simulations of logistic systems and the domain-specific context. Additionally, the framework built on top of the finetuned language model can provide the “NLP shortcut” for simulation modeling. Namely, the framework can produce simple but functionally valid simulations of queuing and inventory control systems given the verbal description in English.
The obtained results, along with the rapid improvement of language models, open the door for significant simplification of the workflow behind the simulation models’ development.

\subsection{Managerial implications}
The presented study highlights the potential of NLP technology, specifically the GPT-3 Codex, to automate the development of simulation models for systems that are critical to logistics operations. As a result of these findings, there are several important managerial implications to consider.

First, this automation can lead to increased efficiency and cost savings for organizations. By automating the development of simulation models, businesses can reduce the time and resources needed to create and maintain these models, as well as the likelihood of human error. Managers should invest in implementing such NLP-based solutions to streamline their logistics operations, ensuring that they remain competitive in the market.

Second, the use of NLP in developing simulation models allows for better utilization of expert knowledge within organizations. By automating the low-level and technical aspects of the modeling process, experts can focus on high-level decision-making and strategic planning. Managers should identify opportunities for reallocating their experts to more critical tasks, thus fostering a culture of innovation and continuous improvement. Additionally, there is the potential for improved communication and collaboration within organizations. Since the GPT-3 Codex can understand domain-specific vocabulary and generate valid simulations based on verbal descriptions, non-technical staff can more easily communicate their needs and requirements to the technical team. Managers should encourage cross-functional collaboration, using NLP technology as a bridge between different departments.

Third, this partial automation of simulation modeling can facilitate better decision-making by providing more accurate and timely models. Managers can use these models to test various scenarios and optimize their logistics strategies, leading to better-informed decisions and more effective resource allocation. Organizations should invest in training their workforce to harness the full potential of these models and be prepared to adapt to the changing business environment.

Finally, the rapid improvement of language models like GPT-3 Codex indicates that the possibilities for further applications and integrations are immense. Managers should stay up-to-date with the latest advancements in NLP technology and explore opportunities to leverage these tools for enhanced business performance. By adopting these technologies early, companies can establish themselves as industry leaders and gain a competitive edge.

In conclusion, the application of NLP for automating the development of simulation models in logistics holds significant potential for increased efficiency, cost savings, better utilization of expert knowledge, improved communication and collaboration, and enhanced decision-making. Managers should actively explore the opportunities provided by these advancements and integrate them into their operations to remain competitive and drive business growth.

\subsection{Future research}
While the work on this paper is ongoing, a new generation of language models is upcoming. GPT-4 continues the trend of increasing the number of parameters in the model as well as the window size (the ability to process more text as input). Additionally, GPT-4 is capable of processing images as well as text inputs, which increases the range of potential applications, including ones related to logistics and supply chain management \citep{openai2023gpt4}. Additionally, the latest developments in the field, including RETRO Transformer by DeepMind \citep{borgeaud_improving_2021} and WebGPT by OpenAI \citep{nakano_webgpt_2021}, successfully demonstrated that the performance close to GPT-3 can be achieved by significantly smaller language models by augmenting them with a way to search and query for information. Adapting the proposed framework to the new generation of language models can be considered a promising direction for future research.

\section*{CRediT authorship contribution statement}
Ilya Jackson: Methodology, Data curation, Software, Validation, Formal analysis, Writing - original draft, Writing - review and editing. Maria Jesus Saenz: Conceptualization, Methodology, Writing - review and editing.

\section*{Acknowledgements}
Our gratitude goes to the OpenAI team for providing early-beta access to GPT-3 Codex.

\section*{Conflict of Interest}
The authors have no conflict of interest to declare.

\bibliography{references}

\end{document}